\newcommand{\myname}{Geoff Boeing}
\newcommand{\myemail}{gboeing@berkeley.edu}
\newcommand{\myaffiliation}{Department of City and Regional Planning\\University of California, Berkeley}
\newcommand{\paperdate}{March 2018}
\newcommand{\papertitle}{Clustering to Reduce Spatial Data Set Size}
\newcommand{\paperkeywords}{Data Science, Machine Learning, DBSCAN, Clustering, Transportation, GIS, Spatial Analysis, Geospatial}

\RequirePackage[l2tabu,orthodox]{nag}   
\documentclass[12pt,onecolumn]{article} 

\usepackage[T1]{fontenc}                
\usepackage[utf8]{inputenc}             

\usepackage[strict,autostyle]{csquotes} 
\usepackage[USenglish]{babel}           
\usepackage{microtype}                  

\usepackage{abstract}                   
\usepackage{authblk}                    
\usepackage{booktabs}                   
\usepackage{caption}                    
\usepackage[final]{draftwatermark}      
\usepackage{endnotes}                   
\usepackage{geometry}                   
\usepackage{graphicx}                   
\usepackage{hyperref}                   
\usepackage{listings}
\usepackage{natbib}                     
\usepackage{rotating}                   
\usepackage{setspace}                   
\usepackage{titlesec}                   
\usepackage{url}                        

\graphicspath{{./figures/}}

\hypersetup{
	pdfauthor={\myname},
	pdftitle={\papertitle},
	pdfsubject={\papertitle},
	pdfkeywords={\paperkeywords},
	pdffitwindow=true,         
	breaklinks=true,           
	colorlinks=false,          
	pdfborder={0 0 0}          
}

\SetWatermarkText{DRAFT}
\SetWatermarkScale{1.3}
\SetWatermarkLightness{0.9}

\begin{document}
	
\title{\papertitle}
\date{\paperdate}
\author[]{\myname \thanks{Email: \href{mailto:\myemail}{\myemail}}}
\affil[]{\myaffiliation}
\maketitle

\section{Introduction}

Traditionally it had been a problem that researchers did not have access to enough spatial data to answer pressing research questions or build compelling visualizations. Today, however, the problem is often that we have too much data. Spatially redundant or approximately redundant points may refer to a single feature (plus noise) rather than many distinct spatial features. We can use density-based clustering to compress such spatial data into a set of representative features.

This paper demonstrates how to reduce the size of a spatial data set of GPS latitude-longitude coordinates using the Python programming language and its scikit-learn implementation of the DBSCAN density-based clustering algorithm. DBSCAN works very well in low-dimension space, such as the two-dimensional feature space in this geospatial example. All of the code discussed here is available in a public repository\endnote{See the code and data repository at \url{https://github.com/gboeing/urban-data-science}} along with the data. 

\section{Data}

How can we reduce the size of a data set down to a smaller set of representative points? Consider a simplified example for the purposes of demonstration: a spatial data set with 1,759 latitude-longitude coordinates. This manageable data set serves merely as a useful object for this demonstration (a related project\endnote{See project clustering millions of points at \url{https://github.com/gboeing/data-visualization}} shows a more complex example of clustering millions of GPS coordinates in-memory).

\begin{figure*}[htbp]
	\centering
	\includegraphics[width=0.8\textwidth]{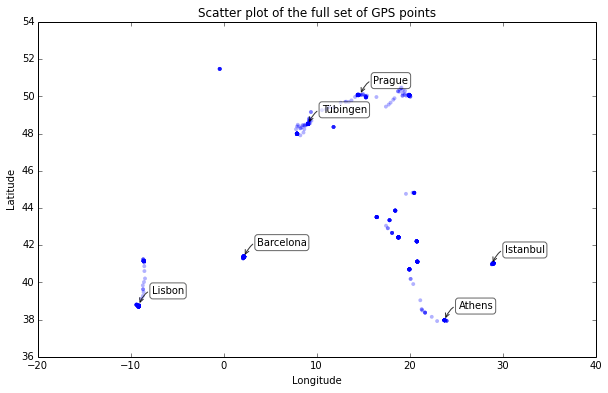}
	\caption{Simple scatter plot of the spatial data set. Labels identify cities with many redundant observations.}
	\label{fig:simple_scatter}
\end{figure*}

These data have been reverse-geocoded to add city and country information. Figure \ref{fig:simple_scatter} shows a simple scatter plot of all the coordinates in the full data set.

\section{Density-based clustering of spatial data}

At the scale of Figure \ref{fig:simple_scatter}, only a few dozen of the 1,759 data points are really visible. Even zoomed-in very close, several locations have hundreds of data points stacked directly on top of each other due to the duration of time spent at one location. Unless we are interested in time dynamics, we simply do not need all of these spatially redundant points---they just bloat the data set's size and refer in duplicate to a single spatial feature.

\subsection{How much data do we need?}

Examine the tight cluster of points in Figure \ref{fig:simple_scatter} representing Barcelona around the coordinate pair (2.15, 41.37). The GPS coordinates in this data set were recorded every 15 minutes: this tight cluster represents hundreds of rows in the data set corresponding to the coordinates of an apartment where the participant resided for a month.

This high number of observations is useful for representing the duration of time spent at certain locations, rather than in transit. However, it grows less useful if the objective is to represent merely where a participant has been. In that case only a single data point is needed for each spatial location to demonstrate that it has been visited. This reduced-size data set would be far easier to map with an on-the-fly rendering tool like JavaScript. It would also be far easier to reverse-geocode only the spatially representative points rather than the thousands or possibly millions of points in a full data set.

\subsection{Clustering algorithms: \textit{k}-means and DBSCAN}

The \textit{k}-means algorithm is likely the most common clustering algorithm \citep{jain_data_2010}. But for spatial data and a problem like this, the DBSCAN algorithm is superior \citep{ester_density-based_1996,zhou_approaches_2000,borah_improved_2004}. Let us consider why.

The \textit{k}-means algorithm groups \textit{N} observations (i.e., rows in an array of coordinates) into \textit{k} clusters. However, \textit{k}-means is not an ideal algorithm for latitude-longitude spatial data because it minimizes variance, not geodetic distance. There is thus substantial distortion at latitudes far from the equator, like those of this data set. The algorithm would still \enquote{work} but its results would be poor and there is not much that can be done to improve them.

With \textit{k}-means, locations where a participant spent a lot of time---such as the Barcelona example above---would still be over-represented because the initial random selection to seed the \textit{k}-means algorithm would select them multiple times. Thus, more rows near a given location in the data set means a higher probability of having more rows selected randomly for that location. Even worse, due to the random seed, many locations would be missing from any clusters, and increasing the number of clusters would still leave patchy gaps throughout the cluster-reduced data set.

Instead, we can use an algorithm that works better with arbitrary distances: scikit-learn's implementation of the DBSCAN algorithm. DBSCAN clusters a spatial data set based on two parameters: a physical distance from each point, and a minimum cluster size. This method works much better for spatial latitude-longitude data.

\subsection{Spatial data clustering with DBSCAN}

We demonstrate clustering with DBSCAN using some simple code examples. First we begin by importing the necessary Python modules and loading the full data set. We convert the latitude and longitude coordinates' columns into a two-dimensional array, called \textit{coords}:

\vspace{1em}
\begin{lstlisting}[language=Python]
import pandas as pd
import numpy as np
import matplotlib.pyplot as plt
from sklearn.cluster import DBSCAN
from geopy.distance import great_circle
from shapely.geometry import MultiPoint
df = pd.read_csv('full-dataset.csv')
coords = df.as_matrix(columns=['lat', 'lon'])
\end{lstlisting}
\vspace{1em}

Next we compute DBSCAN. The $\epsilon$ parameter represents the maximum distance (1.5 kilometers in this example) that points can be from each other to be considered a cluster. The min\_samples parameter is the minimum cluster size allowed---everything else gets classified as noise. We set min\_samples to 1 so that every data point gets assigned to either a cluster or forms its own cluster of size 1, essentially making this a single-link hierarchical clustering process. Nothing will be classified as noise in this example so that we retain far-flung observations as important, non-noise points.

We use the haversine metric with a ball tree space-partitioning data structure to calculate great-circle distances between the points \citep{bhatia_survey_2010}. The $\epsilon$ parameter and the coordinates must be converted to radians, because scikit-learn's haversine metric expects radian units:

\vspace{1em}
\begin{lstlisting}[language=Python]
kms_per_radian = 6371.0088
epsilon = 1.5 / kms_per_radian
db = DBSCAN(eps=epsilon, 
            min_samples=1,
            algorithm='ball_tree',
            metric='haversine').fit(np.radians(coords))
cluster_labels = db.labels_
num_clusters = len(set(cluster_labels))
clusters = pd.Series([coords[cluster_labels == n] for n in range(num_clusters)])
print('Number of clusters: {}'.format(num_clusters))
# Number of clusters: 138
\end{lstlisting}
\vspace{1em}

The fitted DBSCAN model has identified 138 clusters spatial clusters in our data. Unlike \textit{k}-means, DBSCAN does not require us to specify the number of clusters in advance---it determines them automatically based on the $\epsilon$ and min\_samples parameters.

\subsection{Identifying a cluster's centermost point}

Our mission is not complete merely upon identifying clusters of points. To reduce the data set size, we want to identify the coordinates of one point from each cluster that was formed. We could easily just select a random point from each cluster, but it would be more spatially representative if we instead select the point nearest to the cluster's centroid. Note that with DBSCAN, clusters may be non-convex and centroids may fall outside the cluster---however, for this example we simply want to reduce each cluster down to a single point. The point nearest its center is suitable for this use case.

The following function returns the centermost point from a cluster by taking a set of points (i.e., those within the cluster) and returning the point within it that is nearest to some reference point (in this case, the cluster's centroid):

\vspace{1em}
\begin{lstlisting}[language=Python]
def get_centermost_point(cluster):
    centroid = (MultiPoint(cluster).centroid.x, MultiPoint(cluster).centroid.y)
    centermost_point = min(cluster, key=lambda point: great_circle(point, centroid).m)
    return tuple(centermost_point)
centermost_points = clusters.map(get_centermost_point)
\end{lstlisting}
\vspace{1em}

The function above first calculates the centroid's coordinates. Then we find the member of the cluster with the smallest distance to that centroid. Specifically, we do this with a lambda function that calculates each point's distance to the centroid in meters, via the great-circle function. Finally, we return the coordinates of the point that was the least distance from the centroid.

To use this function, we map it to the vector of clusters. In other words, for each element (i.e., cluster) in the vector, the function identifies the centermost point and then assembles all these centermost points into a new vector called centermost\_points. Then we turn these centermost points into a dataframe of points which are spatially representative of the clusters (and in turn, the original full data set):

\vspace{1em}
\begin{lstlisting}[language=Python]
lats, lons = zip(*centermost_points)
rep_points = pd.DataFrame({'lon':lons, 'lat':lats})
\end{lstlisting}
\vspace{1em}

Now we have a set of 138 spatially representative points: one for each cluster. But, we also want the city, country, and timestamp information that was contained in the original full data set. So, for each row of representative points, we pull the full row from the original data set where the latitude and longitude columns match the representative point's latitude and longitude:

\vspace{1em}
\begin{lstlisting}[language=Python]
rs = rep_points.apply(lambda row: df[(df['lat']==row['lat']) & (df['lon']==row['lon'])].iloc[0], axis=1)
\end{lstlisting}
\vspace{1em}

\begin{figure*}[htbp]
	\centering
	\includegraphics[width=0.8\textwidth]{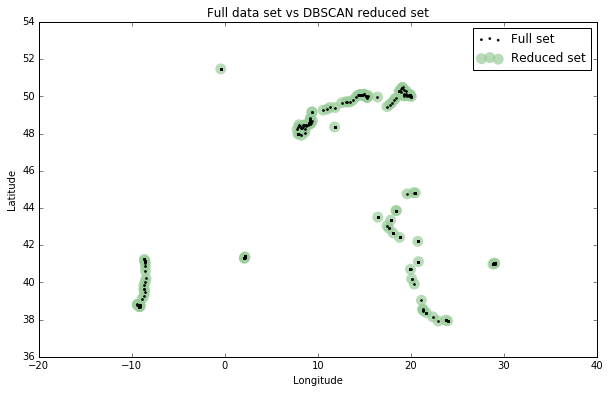}
	\caption{Scatter plot of the clustered spatial data set.}
	\label{fig:clustered_scatter}
\end{figure*}

Now we are all done. We have reduced the original data set down to a spatially representative set of points with full details. We plot the final reduced set of data points versus the original full set in Figure \ref{fig:clustered_scatter} to see how they compare to Figure \ref{fig:simple_scatter}.

\section{Conclusion}

The results in Figure \ref{fig:clustered_scatter} confirm the method's success. We can see the 138 representative points, in green, approximating the spatial distribution of the 1,759 points of the full data set, in black. DBSCAN compressed the data by 92.2\%, from 1,759 points to 138 points. There are no gaps in the reduced data set and heavily-trafficked spots (like Barcelona) are no longer drastically over-represented. This density-based clustering technique has compressed our spatial data into a much smaller set of spatially-representative features.

\IfFileExists{\jobname.ent}{\theendnotes}{}

\setlength{\bibsep}{0.00cm plus 0.05cm} 
\bibliographystyle{apalike}
\bibliography{references}

\end{document}